\documentclass[conference]{IEEEtran}
\IEEEoverridecommandlockouts

\usepackage{cite}
\usepackage{amsmath,amssymb,amsfonts}
\usepackage{algorithmic}
\usepackage{graphicx}
\usepackage{textcomp}
\usepackage{xcolor}
\usepackage{url}
\usepackage{booktabs}
\usepackage{nicefrac}
\usepackage{microtype}
\usepackage{float}
\usepackage{amsthm}
\usepackage[colorlinks=true, linkcolor=blue, citecolor=blue, urlcolor=blue]{hyperref}

\title{Time-Continuous Modeling for Temporal Affective Pattern Recognition in LLM's}

\author{\IEEEauthorblockN{Rezky M. Kam}
\IEEEauthorblockA{\textit{School of Computer Science} \\
\textit{Bina Nusantara University}\\
East Jakarta, Indonesia \\
rezky.kam@binus.ac.id}
\and
\IEEEauthorblockN{Coddy N. Siswanto}
\IEEEauthorblockA{\textit{School of Computer Science} \\
\textit{Bina Nusantara University}\\
East Jakarta, Indonesia \\
coddy.siswanto@binus.ac.id}
}

\begin{document}
\maketitle

\begin{abstract}
\hspace{1.5em}
Text modality decoder models relies on discrete token generation,
where forward attention weights influences individual token representations to predict
the most probable output according to learned patterns
given during backpropagation. However, solely relying on this
approach often lacks a true understanding and introduces black-box interpretation of affective dynamics
in conversations, as true affective understanding involves simultaneous dynamics of
evolving affective trajectories across turn based dialogues. Moreover, due to variability in training data including vast experts in MoE based decoder, it is otherwise not quantifiable to determine which affective personality it belongs through heuristic measure. Here, the mainstream application of online
reinforcement learning is otherwise a challenge for
resource limited environments. Furthermore, we introduce a hybrid encoder-decoder architecture by utilizing decoders with progressive steering of affective trajectory, specifically a
time-aware architecture based on time-continuous patterns and its correlations
with variability of focused group domain affective states by harnessing physics informed neural network, allowing for temporal and longitudinal adaptation. 
Our approach allow machines to mimic the psychological plausibility from its user
over interactions while preserving independent expressiveness, computational efficiency and interpretation.
\end{abstract}

\section{Introduction}
Decoder text generation models can recognize complex semantic meaning and emotional context in single conversational exchanges by picking the highest probability of tokens~\cite{christ2024modelingemotionaltrajectorieswritten, vaswani2017attention}. Though in broader technical perspectives, these models are treated as black-boxes due to the opacity of their internal mechanisms, particularly in how they process the variability of affective states trained across large and diverse corpora. A straightforward alternative to decoder fine-tuning is prompt-based learning, such as Zero-shot or Few-shot prompting using a fine-tuned encoder~\cite{2023, ma2023fairnessguidedfewshotpromptinglarge}. In these approaches, the output of the encoder is typically feature-engineered either through raw logits or temperature-scaled logits in multi-label sentiment extraction tasks:

\[
 p = \{e_i: \tilde{z} | i \in [0, n-1] \} 
\]
\[
 p(e_i) = \frac{\exp(\tilde{z}_i)}{\sum_{j=0}^{n-1} \exp(\tilde{z}_j)} = \frac{\exp(z_i/\delta)}{\sum_{j=0}^{n-1} \exp(z_j/\delta)}
\]

Here, the scale factor $\delta$ modulates the final logit distribution $p$ after applying softmax. As $\delta \rightarrow 0$, the distribution approaches a one-hot encoding, often used to extract the most confident prediction:
\[
 \text{Prompt}(\max(p(e_i))) \rightarrow \text{Predict}(\text{tokens}(p(e_i)))
\]

In contrast, when $\delta \rightarrow \infty$, the distribution becomes uniform, and lower-confidence outputs may be leveraged to introduce diversity or nuanced understanding:
\[
 \text{Prompt}(p(e_{i \neq i^*})) \rightarrow \text{Predict}(\text{tokens}(p(e_{i \neq i^*})))
\]
These strategies are commonly employed in domain-specific chatbots, particularly in applications involving companionship, mental health support, and affective computing. However, a fundamental limitation arises from the discrete and static nature of token-based generation. Human emotions, in contrast, are inherently continuous and evolve over time through dynamic affective trajectories~\cite{conwaysmith2024computationalmechanismsdetachedmindfulness}. As such, current language models struggle to capture the temporal and psychological realism of affective processes, resulting in limited adaptability and interpretability.

To address this limitation, we propose an encoder-decoder framework that integrates the temporal sensitivity of continuous-time modeling with the representational power of large language models and also $\texttt{CEmoFlow}$ dataset consists of emotion soft-labels, timestamp and delay per-utterances. Rather than enhancing emotional understanding solely through prompt engineering or data-driven fine-tuning, we directly intercept the forward pass of a text modality decoder using In-Context Vectors (ICVs)~\cite{liu2024incontextvectorsmakingcontext}, modulated by a time-aware differential system. Specifically, we leverage continuous-time recurrent neural networks (CTRNNs)~\cite{FUNAHASHI1993801} and neural ordinary differential equations (Neural ODEs)~\cite{chen2018neuralode}, trained on continuous affective datasets~\cite{NEURIPS2020_4a5876b4}, to steer token generation along psychologically plausible affective trajectories.

This architecture enables the decoder to account for fine-grained, longitudinal shifts in affective state during multi-turn interactions. Furthermore, by integrating physics-informed neural networks (PINNs) into the affective modeling pipeline, we incorporate domain knowledge to regularize the temporal dynamics, enhancing generalization and interpretability. Importantly, our method retains the computational efficiency of autoregressive generation while augmenting the affective fidelity of the generated dialogue. 

In doing so, we bridge the gap between symbolic discrete modeling and sub-symbolic continuous affect representation, enabling more expressive, emotionally coherent, and user-aligned conversational agents. This contribution has implications not only for affective AI and HCI, but also for broader efforts in explainable and psychologically grounded artificial intelligence.

\begin{figure}[H]
    \centering
    \includegraphics[width=1.0\linewidth]{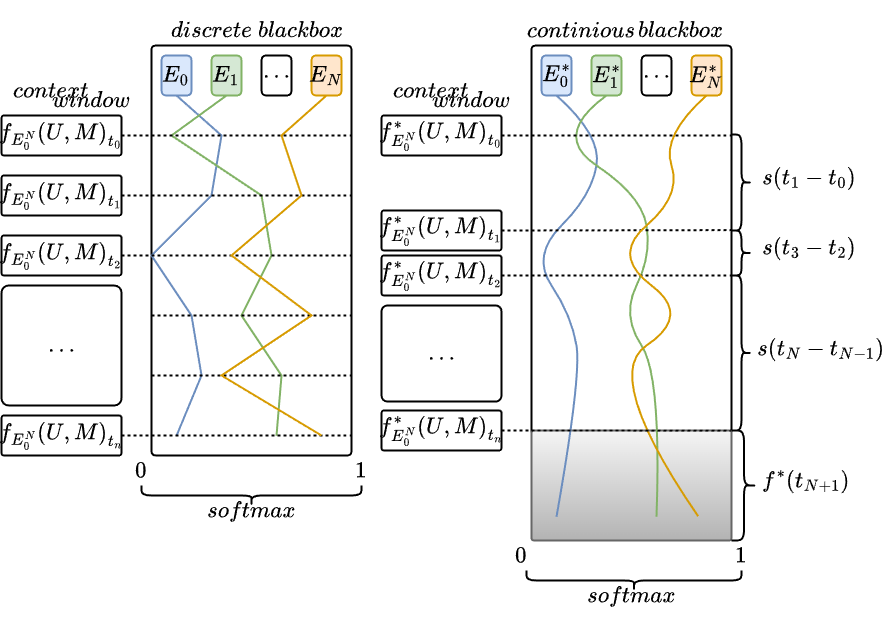}
    \caption{Let $E$ of both functions $f = encoder_{out(E_{0}^{N})}$ presented in discrete fashion and $f^* = encoder \rightarrow LatentODE_{out({E}_{0}^{N})}$ in \textit{continuous} fashion are output probabilities from categories of emotions $N$. The evolving emotions $E$ are interdependent on its own due to softmax constraint, however $f^*$ evolutions are determined by $s$ as delay in seconds that is directly affected by learned differential slopes $\frac{dh}{dt}(f^*(t_{-n}))$ solved with $DOPRI5$ ~\cite{DORMAND198019} during forward pass, further serving as a mathematical form of affective sensing in general context.}
    \label{fig:discvscont}
\end{figure}
Thus, we present a sophisticated emotionally intelligent system by understanding the affective trajectory through real-time interaction and not only through discrete patterns but also through differential slopes of the pattern determined by user delay $s$ between conversational exchanges. With the following structure \textbf{(right diagram)} :
\[
    input_{tok}  \rightarrow f^*\rightarrow ICV(f_{ \ embedding \ })_{\frac{dh}{dt}} 
\]    
\[    
    \rightarrow Decoder \rightarrow output_{tok}
\]
where the original prompt engineering style follows \textbf{(left diagram)} :
\[
    input_{tok} \rightarrow f \rightarrow CraftPrompt \ f({\tilde{E})}_{ \ t} 
\]        
\[
    \rightarrow ContextWindow_{ \ t} \rightarrow output_{tok}
\]
and fine-tuned decoder in practice were inherently :
\[
    input_{tok} \rightarrow FineTunedDecoder \rightarrow output_{tok}
\]

\subsection{Concept}
\begin{center}
\textit{Our approach aims to model the continuous flow of emotional states in human conversation using both temporal encoding and affective shift tracking, in return for a psychologically grounded modeling of a dialogue.}
\end{center}
We propose an architecture that introduces implicit affective intelligence by modeling the dynamic, evolving nature of emotional states in conversation. This capability supports applications in both \textit{Digital Construct} systems, such as virtual agents and emotionally responsive interfaces, and \textit{Companionship} applications, including long-term human-AI interaction in caregiving or therapeutic settings. The model emphasizes the temporality and co-regulation of affective states, enabling a more natural and psychologically informed framework for human-computer dialogue.
The design draws heavily from established theories in cognitive psychology. The \textit{Chameleon Effect} (\textit{Chartrand and Bargh, 1999}) describes the unconscious mimicry of another's affective expressions, a phenomenon that facilitates rapport and empathy in social interaction~\cite{lakin2003chameleon}. \textit{Social Learning Theory} (\textit{Bandura, 1977}) expands this foundation by highlighting emotional learning through observation, and introduces the concept of self-efficacy as a developmental component of emotional intelligence~\cite{JENSEN201568}. These theoretical perspectives inform the affective underpinnings of our system and motivate the tracking of implicit emotional signals across dialogue.
Neuroscientific models provide further structural inspiration. The \textit{Neurocognitive Model of Emotional Contagion} (\textit{Prochazkova and Kret, 2017}) suggests that emotional resonance is grounded in shared neural activations, particularly through sensory mimicry and empathy~\cite{PROCHAZKOVA201799}. This aligns with the \textit{Perception-Action Model} (\textit{Preston and de Waal, 2002}), which posits a direct mapping between the perception of emotion in others and the activation of corresponding internal states~\cite{preston2002empathy}. These affective feedback mechanisms, shown to exist in both in-person and mediated communication, are further supported by recent work such as \textit{Quantification of Interdependent Emotion Dynamics in Online Interactions} (\textit{Luo et al., 2024}), which quantitatively confirms the presence of emotional contagion phenomena in digital platforms~\cite{luo2024quantificationselfexcitedemotiondynamics}. This builds upon earlier empirical studies, including the \textit{Facebook Emotional Contagion} experiment (\textit{2014})~\cite{doi:10.1073/pnas.1320040111}, which demonstrated that subtle shifts in emotional tone within online environments can propagate through social networks.
From a computational standpoint, the architecture is primarily informed by time-sensitive neural modeling. \textit{Liquid Time-Constant Networks} (\textit{Hasani et al., 2020}) introduce neuron-level temporal dynamics through learnable decay constants ($\tau$), inspired by the neurobiology of \textit{C. elegans}, and provide a biologically grounded mechanism for encoding evolving affective states~\cite{Hasani_Lechner_Amini_Rus_Grosu_2021}. Alternatively, \textit{Spiking Neural Networks} (\textit{Maass, 1997}) simulate discrete neuronal events to model symbolic reasoning, though their implementation typically requires specialized neuromorphic hardware~\cite{MAASS19971659}. These temporal modeling paradigms support the affective modulation required for fluid, context-sensitive dialogue understanding.
By combining affective psychology with time-dynamic neural architectures, the proposed model seeks to emulate the emotional fluidity and social alignment present in human conversation. Rather than responding to static emotional cues, it tracks shifts, transitions, and the inertia of affective flow over time. This enables nuanced co-regulation and contextual sensitivity, key features of emotionally intelligent interaction, with significant potential for application in mental health technologies, adaptive conversational agents, and affect-aware robotics.

\section{Continuous NLP}
\textit{\textbf{Preliminaries}\label{noiseclean}} All dataset preprocessing follows the standard and domain-specific practice in deep learning-based NLP, as explained in both literature \textit{chitchat}~\cite{myers2020conversational} and \textit{ emotional dataset for emotional recognition tasks}~\cite{saravia-etal-2018-carer}, however we've made additional steps just in case: \textit{removal of web-links and removal of single character utterances}; resulting in $\mathbf{16}$ rows removal for \textit{chitchat} which afterwards we kept the corresponding conversation windows to those rows as noise, while \textit{emotional dataset for emotional recognition tasks} resulted in $\mathbf{0}$ removed rows.

\subsection{Continuous Dataset} \label{contmodel}  
To effectively model the continuous nature of affective trajectories in conversational exchanges, we selected an open-domain dataset—\textit{chitchat}~\cite{myers2020conversational}—collected as part of a university-hosted competition. The dataset contains $\approx$ 258,145 utterances representing exactly two genuine student interaction per conversation window defined as $t_{segment}$ in \ref{cubicinter}. After noise cleaning \ref{noiseclean}, the dataset has already been chronologically sorted by timestamp, preserving the natural order of the real world conversation flow.

\subsection{Sentiment Annotation}  
Let each row be denoted as $t$, where \( t_{n} \text{and} \ n \approx 258{,}145 \)  . Each row is annotated with soft labels ${E} \in  \{e_0, e_1, \dots, e_N\}$, representing affective states. In our case, we define it as:  
\[
\ E \in \{\text{sadness}_0, \text{joy}_1, \text{love}_2, \text{anger}_3, \text{fear}_4, \text{surprise}_5\}, \quad N=5
\]  
Then let us define a function of an encoder model $f(\tilde {x_t}\forall{x \in \mathcal{\{A\}}})$ that is a fine-tuned ModernBERT~\cite{warner2024smarterbetterfasterlonger} for all utterances $\tilde{x}$ containing only strings $\mathcal{A}$. ModernBERT is chosen due to its generalization performance from RoPE as introduced in RoFormers~\cite{su2023roformerenhancedtransformerrotary} compared to the predecessor in exchange for few more parameters.
Thus, let us denotes $n$ rows that consists of $j$ columns $\forall{e_j}\in {E} \land \ \forall {E}\in \mathbb{R}^{t_n \times 1}$ with mixed precision ~\cite{micikevicius2018mixed} $float$ $\mathcal{F}$;
\begin{equation}
\therefore
\underbrace{{{e_{j}\in \mathcal[0,1]_{\mathcal{F}}\forall{j}\in [0, N]}\land{E_t}\in [0,1]_{\mathcal{F}}}}_{softmax(E)}
\land 
f(\tilde {x_t}\forall{x \in \mathcal{\{A\}}}) 
\end{equation}
\begin{equation}
:=\sum_{t=0}^{t_n}\sum_{j=0}^{e_N}\{e_{tj}|j=0, \cdots, N\}_t
\end{equation}

Before annotation as defined above, the model were fine-tuned based on the dataset—\textit{Emotion Dataset for Emotion Recognition Tasks}~\cite{saravia-etal-2018-carer} preprocessed and curated from English Twitter messages annotated exactly as ${E}$ thereby; the post-training metrics at exactly $step=3200$ as follows:
\begin{figure}[h!]
    \centering
    \includegraphics[width=1.0\linewidth]{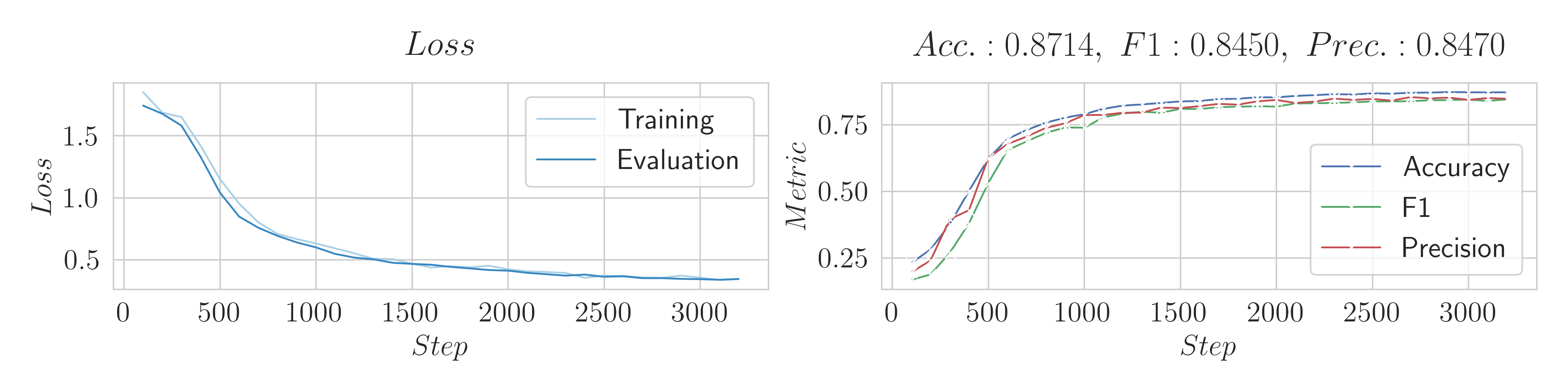}
    \caption{$optim:Lion,lr : 5e-6, lr_{scheduler}:cosine, \ batch : 32, \ \omega \ decay : 1e-2 \ \forall{\theta}, \ bf16 : 1 $. The early stopping is defined as $\forall t \in [T - 5, T] \quad \left| \mathcal{L}(t) - \mathcal{L}_{\text{min}} \right| \leq 1e-2$ noting that $5$ were steps $\neq$ epochs.}
    \label{fig:mcmets}
\end{figure}
\begin{figure}[h!]
    \centering
    \includegraphics[width=0.65\linewidth]{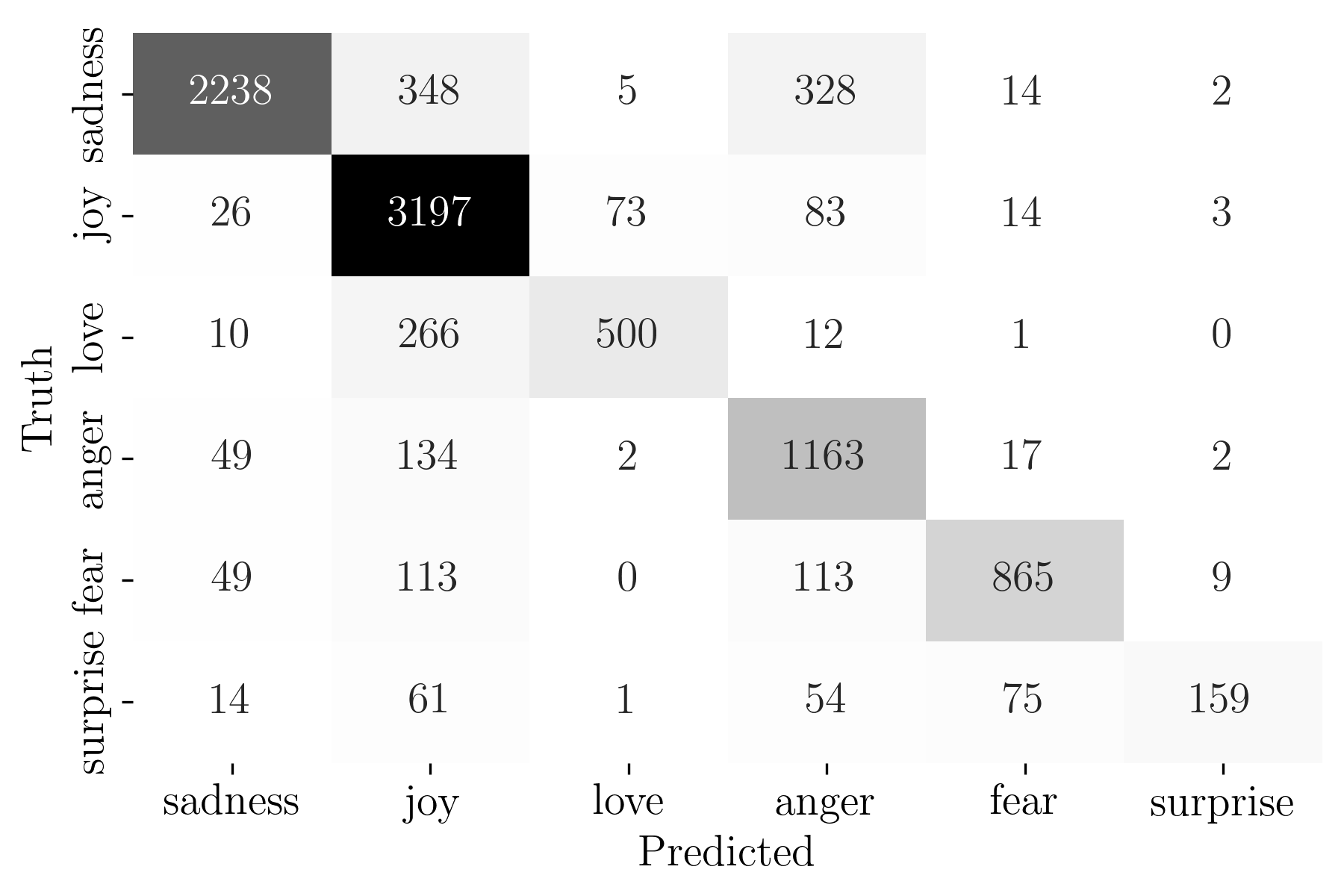}
    \caption{$surprise$ label indicates the least prediction of all due to class imbalances from the dataset~\cite{saravia-etal-2018-carer}, however we've mitigated with generative paraphrasing augmentation via $Qwen2.5$~\cite{qwen2025qwen25technicalreport} for each label $\{love,anger,fear,surprise\}$, resulted in $\texttt{2x}$ more examples but still imbalanced.}
    \label{fig:mcmets}
\end{figure}

\subsection{Cyclic Transform} \label{cyclict}
Cyclical transform from the original timestamp on the annotated—\textit{chitchat} dataset formatted as $\texttt{yyyy:MM:dd:hh:mm:ss}$ is crucial to avoid ambiguity of time formatting for temporal modeling by taking only $\texttt{{hh:mm:ss}}$, then we can define a trigonometric function where $\mathcal{R}$ are radians denoted as ${\forall \mathcal{R}\in  \mathbb{R}^{t_n \times j}\land \forall{j} \in \{sin, cos\}}$, thus for each row $\mathcal{R}$ defined as;

\begin{equation}
\therefore 
\mathcal{R}\in \{sin,cos\}_t 
\end{equation}
\begin{equation}
:= {\sum_{t=0}^{t_n}\sum_{j=0}^{j=1}}((\frac{2\pi \cdot hh_t}{24})_{j},(\frac{2\pi \cdot mm_t}{60})_{j}, (\frac{2\pi \cdot ss_t}{60})_{j})_t
\end{equation}

Therefore, let us denote delay column $\tau$ as $\tau \in \{ss_t-ss_{t-1}\}\forall{t\neq0} \land \ \forall{\tau} \in \mathbb{R}^{t_n \times 1}$, precisely noting that $ss$ should \textit{not} be transformed cyclically \textbf{before} computing $\tau$ and should be transformed cyclically \textbf{after} computing $\tau$, as $\tau$ should be in absolute form of $ss$ to be continuously structured from $[t_0,t_n]$ for a truly monotonic continuous integration $\text{of } \  \tilde{T}=\int_{t_0}^{t_n}f(t)dt$, thus let us define a sum-cumulative expression of a discrete integration;
\begin{equation}
\therefore \tau_{t_0}:[0]\land\forall \tau_{t \neq 0}\in[1.00,86400.00]\land
\tilde{T}:\int_{t_0}^{t_n}f(t)dt
\end{equation}
\begin{equation}
:=\sum_{t=0}^{t_n}\tau_t|ss_t-ss_{t-1}|
\end{equation}

\subsection{Affective Magnitude} The affective magnitude $\Delta$ refers to the general term of \textit{emotional shifts} by differentiating the affective state at $t$ and $t{-1}$, denoted as $\forall{\Delta}\in \mathbb{R}^{t_n \times 1}$, simply by calculating the \texttt{L1-Norm} that defines each row as;

\begin{equation}
\therefore e_j\Delta_{t_0}\in[0]\land \forall e_{tj}\in{[0,1]}, 
\Delta_t
\end{equation}

\begin{equation}
:=\sum_{t=1}^{t_n} \sum_{j=0}^{e_N} |e_{tj}-e_{tj-1}|
\end{equation}

\subsection{Continuous Interpolation} \label{cubicinter}
Modeling discrete data into continuous trajectory requires us to interpolate for $\tilde{T}=\int_{t_0}^{t_n}f(t)dt$ \textit{Eq.}\ref{cyclict} with $\textbf{\texttt{scipy.interpolate.CubicHermiteSpline}}$~\cite{2020SciPy-NMeth}, noting that $\tilde{T}$ is a synthetic global variable denoted as $\tilde{T}\in \mathbb{C}^{\tilde{t}_n \times 1}$, whereas \textit{Non-Hermite} Cubic Spline Interpolation computes unnecessary outliers from overshooting transition slopes of $S_{i}(t)$ as a result from $C^2$ smoothness that would affect $\tilde{T}$, formally;

\begin{equation}
\tilde{T}_{C^{1 \lor 2}}\approx \sum_{i=0}^{n-1}\int_{t_i}^{t_{i+1}}S_{i}(t)dt
\end{equation}
\begin{equation}
:=[f(t_{i})t_{i+1}-t_{i}
\end{equation}
\begin{equation}
+\frac{f(b_{i})t_{i+1}-t_{i}^{2}}{2}+\frac{f(c_{i})t_{i+1}-t_{i}^{3}}{3}+\frac{f(d_{i})t_{i+1}-t_{i}^{4}}{4}] 
\end{equation}
\begin{equation}
\underbrace{f(c_i)}_{2nd  \ derivative }=(t_{i+1}-t_{i})_{i-1}(c_{i-1}+2((t_{i+1}-t_{i})_{i-1}
\end{equation}
\begin{equation}
+(t_{i+1}-t_{i}))c_{i}+(t_{i+1}-t_{i})c_{i+1}
\end{equation}
\begin{equation}
=3(\frac{f(t_{i+1})-f(t_i)}{t_{i+1}-t_{i}}-\frac{f(t_{i})-f(t_{i-1})}{(t_{i+1}-t_{i})_{i-1}},
\end{equation}
\begin{equation}
\underbrace{f(b_i)}_{1st \ derivative}=\frac{f(t_{i+1})-f(t_i)}{t_{i+1}-t_{i}}-\frac{t_{i+1}-t_{i}}{3}(2c_{i}+c_{i+1})  
\end{equation}
\begin{equation}
\underbrace{f(d_{i)}}_{coefficient}=\frac{c_{i+1}-c_{i}}{3(t_{i+1}-t_{i})}
, \quad \therefore v_i=f(t_i)
\end{equation}

Thus, in our case for \textit{${chitchat}$}, introducing temporal segmentation is crucial even for Cubic Hermit Spline to prevent noise from global outliers by encapsulating slope $\forall t_{C^1} \in[t_{start}, t_{end}]$ and $C^0$ between ${t}_{segment}$, the segmented interpolation computes unique sequence of differential slope patterns that differentiates it from other segments due to localized effect while retaining its continuous nature. Therefore, let us define each segment with implicitly interdependent variables;
\begin{equation}
v={t}_{segment} \in \{{E},{\mathcal{R}}_{sin,cos},\tau,\Delta\}_t
\end{equation}
\begin{equation}
\land \forall \ {t_{segment}}\in[t_{start},t_{end}]
\end{equation}

\begin{equation}
, \quad \text{where} \quad \ 
{E}\in \{e_0, \cdots, e_N\} \quad 
\end{equation}
\begin{equation}
,\quad {\mathcal{R}}\in \{\tilde{h},\tilde{m},\tilde{s}\}_{sin,cos} 
\end{equation}
Thus, we can compute the slope \( S \) by interpolating the differentials of the forward difference at \( S_0 \), the central difference at \( S_i \), and the backward difference at \( S_{n-1} \) defined as;
\begin{equation}
forward_{\ S_0}=\frac{v_1-v_0}{{t}_1-{t}_0}
\end{equation}
\begin{equation}
,central_{\ S}=\frac{v_{i+1}-v_{i-1}}{t_{i+1}-{t}_{i-1}}
\end{equation}
\begin{equation}
,backward_{\ S_{n-1}}=\frac{v_{n}-v_{n-1}}{{t}_{n}-{t}_{n-1}} 
\end{equation}

\begin{equation}
\text{finally, } \ \  \forall \ {{t}_{segment}\in \{t_i,t_{i+1}\} \equiv \tilde{T}_{C^{0\lor 1}}\in \{\tau \}}
\end{equation}

\begin{equation}
\int_{t_0}^{t_n}f(t)=h_{00}(x)\cdot v_{i} \  \text{where} \quad h_{00}(x)=2x^{3}-3x^{2}+1,
\end{equation}
\begin{equation}
+h_{10}(x)\cdot (t_{i+1}-t_{i})\cdot m_{i} \ \text{where} \quad h_{10}(x)=x^{3}-2x^{2}+x, 
\end{equation}
\begin{equation}
+h_{01}(x)\cdot v_{i+1} \ \text{where} \quad h_{01}(x)=-2x^{3}+3x^{2},
\end{equation}
\begin{equation}
+h_{11}(x)\cdot (t_{i+1}-t_{i})\cdot m_{i+1} \ \text{where} \quad h_{11}(x)=x^{3}-x^{2}
\end{equation}
\begin{equation}
, \quad \text{and} \quad x=\frac{t-t_i}{t_{i+1}-t_{i}}
\end{equation}

\subsection{CEmoFlow Dataset}
We introduce the $\texttt{CEmoFlow}$ dataset, the structure consists of chronologically sorted conversational exchanges that its utterances are completely anonymized from \textit{chitchat}~\cite{myers2020conversational}., thereby providing a continuously interpolated emotion soft labels and its timestamp that fosters the use of PINN's for simulating real-world emotional dynamics.
\begin{figure}[H]
    \centering
    \includegraphics[width=1.0\linewidth]{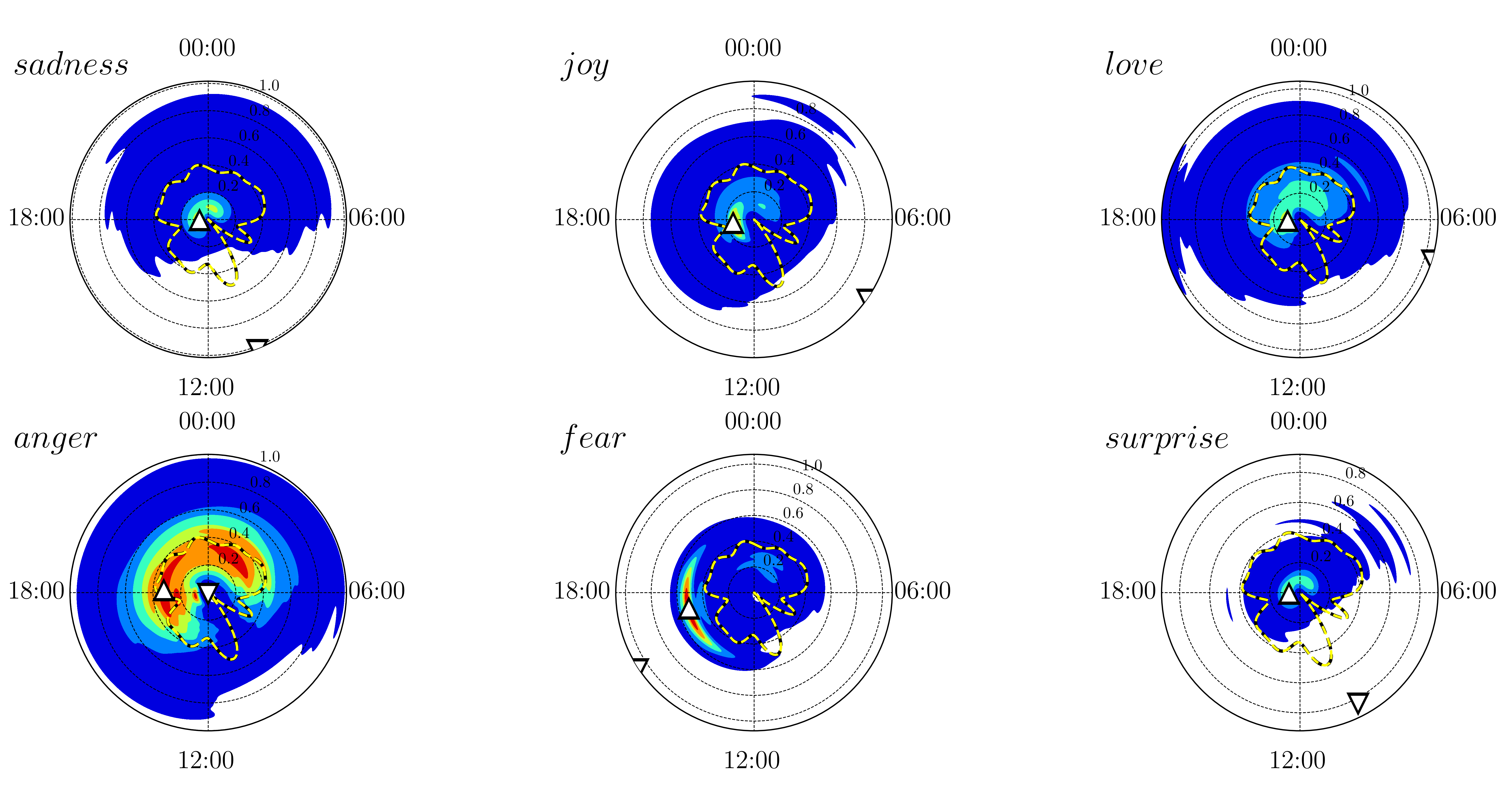}
    \caption{Gaussian KDE with $250.000$ samples where $hh_{sin, cos}$ transforms to $arctan2(hh_{sin, cos})$ that represents each soft labels distribution on $x \in \{\text{\textbf{A}nte \textbf{M}eridiem},\text{\textbf{P}ost \textbf{M}eridiem}\}$ axis following their intensity peak by the contour on $y \in \forall e_N \in  [0,1]$ axis, the yellow line represents $\mu \Delta \in [0,1]$ (normalized), whereas $\bigtriangleup$ marks highest distribution and $\bigtriangledown$ marks lowest distribution density of the contour.}
    \label{fig:radar}
\end{figure}

\begin{table}[ht]
\centering
\scriptsize
\setlength{\tabcolsep}{1.2pt}
\resizebox{\linewidth}{!}{%
\begin{tabular}{lrrrrrrrr}
\toprule
$X$ & $count$ & $mean$ & $std$ & $min$ & $25\%$ &  $50\%$ & $75\%$ & $max$ \\
\midrule
$\tilde{T}$     & 6.76e+06 & 3.16e+06 & 1.89e+06 & 0.00e+00 & 1.64e+06 & 2.99e+06 & 4.78e+06 & 6.61e+06 \\
$\tau$         & 6.76e+06 & 4.37e+01 & 6.62e+01 & -5.35e+01 & 6.93e+00 & 1.94e+01 & 5.00e+01 & 5.59e+02 \\
$e_0$       & 6.76e+06 & 8.82e-02 & 8.71e-02 & -1.18e-01 & 3.97e-02 & 6.77e-02 & 1.08e-01 & 1.17e+00 \\
$e_1$           & 6.76e+06 & 1.57e-01 & 1.24e-01 & -1.25e-01 & 7.51e-02 & 1.26e-01 & 1.99e-01 & 1.09e+00 \\
$e_2$           & 6.76e+06 & 1.77e-01 & 1.36e-01 & -1.22e-01 & 8.59e-02 & 1.37e-01 & 2.28e-01 & 1.10e+00 \\
$e_3$          & 6.76e+06 & 2.89e-01 & 1.47e-01 & -9.60e-02 & 1.84e-01 & 2.70e-01 & 3.69e-01 & 1.14e+00 \\
$e_4$           & 6.76e+06 & 2.06e-01 & 1.36e-01 & -1.20e-01 & 1.04e-01 & 1.81e-01 & 2.81e-01 & 1.13e+00 \\
$e_5$       & 6.76e+06 & 8.30e-02 & 6.96e-02 & -1.16e-01 & 4.30e-02 & 6.96e-02 & 1.02e-01 & 9.84e-01 \\
$\Delta$     & 6.76e+06 & 6.78e-01 & 3.35e-01 & -2.50e-01 & 4.43e-01 & 6.37e-01 & 8.77e-01 & 2.19e+00 \\
$\mathcal{R} \ sin(h)$     & 6.76e+06 & -1.34e-01 & 7.02e-01 & -1.29e+00 & -8.66e-01 & -2.59e-01 & 5.00e-01 & 1.29e+00 \\
$\mathcal{R} \ cos(h)$     & 6.76e+06 & 4.23e-01 & 5.52e-01 & -1.30e+00 & 0.00e+00 & 5.21e-01 & 8.66e-01 & 1.29e+00 \\
$\mathcal{R} \ sin(m)$   & 6.76e+06 & -1.86e-02 & 6.95e-01 & -1.31e+00 & -6.98e-01 & -9.90e-03 & 6.69e-01 & 1.33e+00 \\
$\mathcal{R} \ cos(m)$   & 6.76e+06 & 2.13e-02 & 7.09e-01 & -1.27e+00 & -6.76e-01 & 6.99e-02 & 7.21e-01 & 1.30e+00 \\
$\mathcal{R} \ sin(s)$   & 6.76e+06 & -1.03e-02 & 6.44e-01 & -1.40e+00 & -5.96e-01 & -1.58e-02 & 5.77e-01 & 1.39e+00 \\
$\mathcal{R} \ cos(s)$   & 6.76e+06 & 6.61e-03 & 6.38e-01 & -1.38e+00 & -5.72e-01 & 8.10e-03 & 5.86e-01 & 1.41e+00 \\
\bottomrule
\end{tabular}
}
\label{tab:summary_stats}
\end{table}

\section{Limitation}
The limitations provided by \texttt{CEmoFlow} and the concept of using PINN's for Text Gen LLM's latent space steering~\cite{liu2024incontextvectorsmakingcontext} resulted in significantly higher implementation complexity, experiments are done with Non-Recurrent Multioutput-Regression Neural ODE~\cite{chen2018neuralode} by training it on $y \in \{e_0,e_1,e_2,e_3,e_4,e_5\}$ where $X \subset y$ resulted in higher compute time using $Dopri5$ by $\texttt{121:45:12}$ until convergence with early stopping, whereas, Multioutput-Regression LSTM with the same training configuration resulted in only $\texttt{9:11:56}$ where the only limitation is its black-box nature, the experiment were conducted using $\texttt{RTX4070}$ consisting of $\texttt{4608 CUDA Cores}$ and $\texttt{144 Tensor Cores}$, however solving the ODE are done purely with $\texttt{CPU}$ and shapes the latent space later on the $\texttt{GPU}$.

\bibliography{references}

@misc{vaswani2017attention,
  author    = {Vaswani, A. and Shazeer, N. and Parmar, N. and Uszkoreit, J. and Jones, L. and Gomez, A. N. and Kaiser, L. and Polosukhin, I.},
  title     = {Attention Is All You Need},
  journal   = {Advances in Neural Information Processing Systems},
  volume    = {30},
  pages     = {5998--6008},
  year      = {2017}
}

@misc{christ2024modelingemotionaltrajectorieswritten,
      title={Modeling Emotional Trajectories in Written Stories Utilizing Transformers and Weakly-Supervised Learning}, 
      author={Lukas Christ and Shahin Amiriparian and Manuel Milling and Ilhan Aslan and Björn W. Schuller},
      year={2024},
      eprint={2406.02251},
      archivePrefix={arXiv},
      primaryClass={cs.CL},
      url={https://arxiv.org/abs/2406.02251}, 
}

@misc{luo2024quantificationselfexcitedemotiondynamics,
      title={Quantification of the Self-Excited Emotion Dynamics in Online Interactions}, 
      author={Yishan Luo and Didier Sornette and Sandro Claudio Lera},
      year={2024},
      eprint={2408.05700},
      archivePrefix={arXiv},
      primaryClass={cs.SI},
      url={https://arxiv.org/abs/2408.05700}, 
}

@misc{liu2024incontextvectorsmakingcontext,
      title={In-context Vectors: Making In Context Learning More Effective and Controllable Through Latent Space Steering}, 
      author={Sheng Liu and Haotian Ye and Lei Xing and James Zou},
      year={2024},
      eprint={2311.06668},
      archivePrefix={arXiv},
      primaryClass={cs.LG},
      url={https://arxiv.org/abs/2311.06668}, 
}

@misc{conwaysmith2024computationalmechanismsdetachedmindfulness,
      title={The Computational Mechanisms of Detached Mindfulness}, 
      author={Brendan Conway-Smith and Robert L. West},
      year={2024},
      eprint={2409.15289},
      archivePrefix={arXiv},
      primaryClass={q-bio.NC},
      url={https://arxiv.org/abs/2409.15289}, 
}

@inproceedings{saravia-etal-2018-carer,
    title = "{CARER}: Contextualized Affect Representations for Emotion Recognition",
    author = "Saravia, Elvis  and
      Liu, Hsien-Chi Toby  and
      Huang, Yen-Hao  and
      Wu, Junlin  and
      Chen, Yi-Shin",
    editor = "Riloff, Ellen  and
      Chiang, David  and
      Hockenmaier, Julia  and
      Tsujii, Jun{'}ichi",
    booktitle = "Proceedings of the 2018 Conference on Empirical Methods in Natural Language Processing",
    month = oct # "-" # nov,
    year = "2018",
    address = "Brussels, Belgium",
    publisher = "Association for Computational Linguistics",
    url = "https://aclanthology.org/D18-1404/",
    doi = "10.18653/v1/D18-1404",
    pages = "3687--3697"
}

@article{chen2018neuralode,
  title={Neural Ordinary Differential Equations},
  author={Chen, Ricky T. Q. and Rubanova, Yulia and Bettencourt, Jesse and Duvenaud, David},
  journal={Advances in Neural Information Processing Systems},
  year={2018}
}

@inproceedings{NEURIPS2020_4a5876b4,
 author = {Kidger, Patrick and Morrill, James and Foster, James and Lyons, Terry},
 booktitle = {Advances in Neural Information Processing Systems},
 title = {Neural Controlled Differential Equations for Irregular Time Series},
 volume = {33},
 year = {2020}
}

@article{FUNAHASHI1993801,
    title = {Approximation of dynamical systems by continuous time recurrent neural networks},
    journal = {Neural Networks},
    volume = {6},
    number = {6},
    pages = {801-806},
    year = {1993},
    issn = {0893-6080},
    doi = {https://doi.org/10.1016/S0893-6080(05)80125-X},
    author = {Kenichi Funahashi and Yuichi Nakamura},
}

@incollection{JENSEN201568,
    title = {Delinquency, Sociology of},
    editor = {James D. Wright},
    booktitle = {International Encyclopedia of the Social \& Behavioral Sciences (Second Edition)},
    publisher = {Elsevier},
    edition = {Second Edition},
    address = {Oxford},
    pages = {68-74},
    year = {2015},
    author = {Gary F. Jensen}
}

@article{lakin2003chameleon,
  title={The chameleon effect as social glue: Evidence for the evolutionary significance of nonconscious mimicry},
  author={Lakin, Jessica L. and Jefferis, Valerie E. and Cheng, Clara M. and Chartrand, Tanya L.},
  journal={Journal of Nonverbal Behavior},
  volume={27},
  number={3},
  pages={145--162},
  year={2003},
  publisher={Springer},
  doi={10.1023/A:1025389814290}
}

@article{PROCHAZKOVA201799,
title = {Connecting minds and sharing emotions through mimicry: A neurocognitive model of emotional contagion},
journal = {Neuroscience \& Biobehavioral Reviews},
volume = {80},
pages = {99-114},
year = {2017},
issn = {0149-7634},
doi = {https://doi.org/10.1016/j.neubiorev.2017.05.013},
url = {https://www.sciencedirect.com/science/article/pii/S0149763416306704},
author = {Eliska Prochazkova and Mariska E. Kret},
}

@article{preston2002empathy,
  title={Empathy: Its ultimate and proximate bases},
  author={Preston, Stephanie D and De Waal, Frans BM},
  journal={Behavioral and brain sciences},
  volume={25},
  number={1},
  pages={1--20},
  year={2002},
  publisher={Cambridge University Press}
}

@article{MAASS19971659,
title = {Networks of spiking neurons: The third generation of neural network models},
journal = {Neural Networks},
volume = {10},
number = {9},
pages = {1659-1671},
year = {1997},
issn = {0893-6080},
doi = {https://doi.org/10.1016/S0893-6080(97)00011-7},
url = {https://www.sciencedirect.com/science/article/pii/S0893608097000117},
author = {Wolfgang Maass},
}

@article{DORMAND198019,
title = {A family of embedded Runge-Kutta formulae},
journal = {Journal of Computational and Applied Mathematics},
volume = {6},
number = {1},
pages = {19-26},
year = {1980},
issn = {0377-0427},
doi = {https://doi.org/10.1016/0771-050X(80)90013-3},
url = {https://www.sciencedirect.com/science/article/pii/0771050X80900133},
author = {J.R. Dormand and P.J. Prince}
}

@article{Hasani_Lechner_Amini_Rus_Grosu_2021, title={Liquid Time-constant Networks}, volume={35}, url={https://ojs.aaai.org/index.php/AAAI/article/view/16936}, DOI={10.1609/aaai.v35i9.16936}, abstractNote={We introduce a new class of time-continuous recurrent neural network models. Instead of declaring a learning system’s dynamics by implicit nonlinearities, we construct networks of linear first-order dynamical systems modulated via nonlinear interlinked gates. The resulting models represent dynamical systems with varying (i.e., liquid) time-constants coupled to their hidden state, with outputs being computed by numerical differential equation solvers. These neural networks exhibit stable and bounded behavior, yield superior expressivity within the family of neural ordinary differential equations, and give rise to improved performance on time-series prediction tasks. To demonstrate these properties, we first take a theoretical approach to find bounds over their dynamics, and compute their expressive power by the trajectory length measure in a latent trajectory space. We then conduct a series of time-series prediction experiments to manifest the approximation capability of Liquid Time-Constant Networks (LTCs) compared to classical and modern RNNs.}, number={9}, journal={Proceedings of the AAAI Conference on Artificial Intelligence}, author={Hasani, Ramin and Lechner, Mathias and Amini, Alexander and Rus, Daniela and Grosu, Radu}, year={2021}, month={May}, pages={7657-7666} }

@article{
doi:10.1073/pnas.1320040111,
author = {Adam D. I. Kramer  and Jamie E. Guillory  and Jeffrey T. Hancock },
title = {Experimental evidence of massive-scale emotional contagion through social networks},
journal = {Proceedings of the National Academy of Sciences},
volume = {111},
number = {24},
pages = {8788-8790},
year = {2014},
doi = {10.1073/pnas.1320040111},
URL = {https://www.pnas.org/doi/abs/10.1073/pnas.1320040111},
eprint = {https://www.pnas.org/doi/pdf/10.1073/pnas.1320040111},
}

@article{myers2020conversational,
    title={Conversational Scaffolding: An Analogy-Based Approach to Response Prioritization in Open-Domain Dialogs},
    journal = {Proceedings of the 12th International Conference on Agents and Artificial Intelligence},
    author={Myers, Will and Etchart, Tyler and Fulda, Nancy},
    URL = {https://www.scitepress.org/Papers/2020/89399/89399.pdf},
    year={2020}
}

@misc{warner2024smarterbetterfasterlonger,
      title={Smarter, Better, Faster, Longer: A Modern Bidirectional Encoder for Fast, Memory Efficient, and Long Context Finetuning and Inference}, 
      author={Benjamin Warner and Antoine Chaffin and Benjamin Clavié and Orion Weller and Oskar Hallström and Said Taghadouini and Alexis Gallagher and Raja Biswas and Faisal Ladhak and Tom Aarsen and Nathan Cooper and Griffin Adams and Jeremy Howard and Iacopo Poli},
      year={2024},
      eprint={2412.13663},
      archivePrefix={arXiv},
      primaryClass={cs.CL},
      url={https://arxiv.org/abs/2412.13663}, 
}

@misc{su2023roformerenhancedtransformerrotary,
      title={RoFormer: Enhanced Transformer with Rotary Position Embedding}, 
      author={Jianlin Su and Yu Lu and Shengfeng Pan and Ahmed Murtadha and Bo Wen and Yunfeng Liu},
      year={2023},
      eprint={2104.09864},
      archivePrefix={arXiv},
      primaryClass={cs.CL},
      url={https://arxiv.org/abs/2104.09864}, 
}

@inproceedings{
micikevicius2018mixed,
title={Mixed Precision Training},
author={Paulius Micikevicius and Sharan Narang and Jonah Alben and Gregory Diamos and Erich Elsen and David Garcia and Boris Ginsburg and Michael Houston and Oleksii Kuchaiev and Ganesh Venkatesh and Hao Wu},
booktitle={International Conference on Learning Representations},
year={2018},
url={https://openreview.net/forum?id=r1gs9JgRZ},
}

@ARTICLE{2020SciPy-NMeth,
  author  = {Virtanen, Pauli and Gommers, Ralf and Oliphant, Travis E. and
            Haberland, Matt and Reddy, Tyler and Cournapeau, David and
            Burovski, Evgeni and Peterson, Pearu and Weckesser, Warren and
            Bright, Jonathan and {van der Walt}, St{\'e}fan J. and
            Brett, Matthew and Wilson, Joshua and Millman, K. Jarrod and
            Mayorov, Nikolay and Nelson, Andrew R. J. and Jones, Eric and
            Kern, Robert and Larson, Eric and Carey, C J and
            Polat, {\.I}lhan and Feng, Yu and Moore, Eric W. and
            {VanderPlas}, Jake and Laxalde, Denis and Perktold, Josef and
            Cimrman, Robert and Henriksen, Ian and Quintero, E. A. and
            Harris, Charles R. and Archibald, Anne M. and
            Ribeiro, Ant{\^o}nio H. and Pedregosa, Fabian and
            {van Mulbregt}, Paul and {SciPy 1.0 Contributors}},
  title   = {{{SciPy} 1.0: Fundamental Algorithms for Scientific
            Computing in Python}},
  journal = {Nature Methods},
  year    = {2020},
  volume  = {17},
  pages   = {261--272},
  adsurl  = {https://rdcu.be/b08Wh},
  doi     = {10.1038/s41592-019-0686-2},
}

@inproceedings{2023, series={RANLP},
   title={A Practical Survey on Zero-shot Prompt Design for In-context Learning},
   url={http://dx.doi.org/10.26615/978-954-452-092-2_069},
   DOI={10.26615/978-954-452-092-2_069},
   booktitle={Proceedings of the Conference Recent Advances in Natural Language Processing - Large Language Models for Natural Language Processings},
   publisher={INCOMA Ltd., Shoumen, BULGARIA},
   author={Li, Yinheng},
   year={2023},
   pages={641–647},
   collection={RANLP} }

@misc{ma2023fairnessguidedfewshotpromptinglarge,
      title={Fairness-guided Few-shot Prompting for Large Language Models}, 
      author={Huan Ma and Changqing Zhang and Yatao Bian and Lemao Liu and Zhirui Zhang and Peilin Zhao and Shu Zhang and Huazhu Fu and Qinghua Hu and Bingzhe Wu},
      year={2023},
      eprint={2303.13217},
      archivePrefix={arXiv},
      primaryClass={cs.CL},
      url={https://arxiv.org/abs/2303.13217}, 
}

@misc{qwen2025qwen25technicalreport,
      title={Qwen2.5 Technical Report}, 
      author={Qwen and : and An Yang and Baosong Yang and Beichen Zhang and Binyuan Hui and Bo Zheng and Bowen Yu and Chengyuan Li and Dayiheng Liu and Fei Huang and Haoran Wei and Huan Lin and Jian Yang and Jianhong Tu and Jianwei Zhang and Jianxin Yang and Jiaxi Yang and Jingren Zhou and Junyang Lin and Kai Dang and Keming Lu and Keqin Bao and Kexin Yang and Le Yu and Mei Li and Mingfeng Xue and Pei Zhang and Qin Zhu and Rui Men and Runji Lin and Tianhao Li and Tianyi Tang and Tingyu Xia and Xingzhang Ren and Xuancheng Ren and Yang Fan and Yang Su and Yichang Zhang and Yu Wan and Yuqiong Liu and Zeyu Cui and Zhenru Zhang and Zihan Qiu},
      year={2025},
      eprint={2412.15115},
      archivePrefix={arXiv},
      primaryClass={cs.CL},
      url={https://arxiv.org/abs/2412.15115}, 
}

\end{document}